%% file: main.tex
\documentclass[letterpaper, 10pt, conference]{ieeeconf} 
\IEEEoverridecommandlockouts      
\makeatletter
\let\NAT@parse\undefined
\makeatother

\usepackage{times}
\usepackage{multicol}
\usepackage{xspace}
\usepackage{amssymb,amsfonts}
\usepackage{amsmath}
\usepackage{graphicx}
\usepackage{wrapfig}
\usepackage[caption=false,font=footnotesize]{subfig}
\usepackage{algorithm}
\usepackage[noend]{algorithmic}
\usepackage{cite}
\usepackage{xcolor}
\usepackage{verbatim}

\input{tex/macros}

\begin{document}

\title{ Motion Planning for Multi-Link Robots \\ by 
  Implicit Configuration-Space Tiling*}

\author{%
  Oren Salzman$^{\dagger}$, Kiril Solovey$^{\dagger}$ and Dan
  Halperin$^{\dagger}$
  \thanks{$^{\dagger}$ Oren Salzman, Kiril Solovey and Dan Halperin
    are with the Blavatnik School of Computer Science, Tel-Aviv
    University, Israel. Email:
    \url{[orenzalz,kirilsol,danha]@post.tau.ac.il}}%
  \thanks{* This work has been supported in part by the Israel Science
    Foundation (grant no. 825/15), by the German-Israeli Foundation
    (grant no. 1150-82.6/2011), and by the Hermann Minkowski--Minerva
    Center for Geometry at Tel Aviv University. Kiril Solovey is also
    supported by the Clore Israel Foundation.}
}

\maketitle

\begin{abstract}
  We study the problem of motion-planning for free-flying multi-link
  robots and develop a sampling-based algorithm that is specifically
  tailored for the task.  Our approach exploits the fact that the set
  of configurations for which the robot is \emph{self-collision free}
  is independent of the obstacles or of the exact placement of the
  robot. This allows for decoupling between costly self-collision
  checks on the one hand, which we do off-line (and can even be stored
  permanently on the robot's controller), and collision with obstacles
  on the other hand, which we compute in the query phase. In
  particular, given a specific robot type our algorithm precomputes a
  \emph{tiling roadmap}, which efficiently and implicitly encodes the
  self-collision free \mbox{(sub-)space} over the entire configuration
  space, where the latter can be infinite for that matter. To answer a
  query, which consists of a start position, a target region, and a
  \emph{workspace environment}, we traverse the tiling roadmap while
  only testing for collisions with obstacles. Our algorithm suggests
  more flexibility than the prevailing paradigm in which a precomputed
  roadmap depends both on the robot and on the scenario at hand.  We
  demonstrate the effectiveness of our approach on open and
  closed-chain multi-link robots, where in some settings our algorithm
  is more than fifty times faster than commonly used,  as well as
  state-of-the-art solutions.
\end{abstract}

\IEEEpeerreviewmaketitle

\input{tex/intro}
\input{tex/preliminaries}
\input{tex/tiling_roadmap}
\input{tex/implementation_details}
\input{tex/coverage}

\input{tex/evaluation}
\input{tex/future_work}


\end{document}

%% file: tex/macros.tex
\def\C{\mathcal{C}}

\def\F{\mathcal{F}}

\def\E{\mathcal{E}}
\def\S{\mathcal{S}}
\def\G{\mathcal{G}}

\def\T{\mathcal{T}}

\def\V{\mathcal{V}}

\def\W{\mathcal{W}}
\def\S{\mathcal{S}}

\def\dC{\mathbb{C}}

\def\dN{\mathbb{N}}

\def\dR{\mathbb{R}}

\def\Origin{\mathbf{0}}

\renewcommand{\leq}{\leqslant}

\newcommand{\Cpp}{C\raise.08ex\hbox{\tt ++}\xspace}

\newcommand{\trd}{TR-dRRT\xspace}

\newtheorem{defin}{Definition}
  
\newtheorem{theo}[defin]{Theorem}
  
\newtheorem{lemma}[defin]{Lemma}
  
\newtheorem{propo}[defin]{Proposition}
  
\newtheorem{coro}[defin]{Corollary}
  
\newtheorem{obse}[defin]{Observation}
  \newenvironment{observation}{\begin{obse} \sl}{\end{obse}}
\newtheorem{rem}[defin]{Remark}

\def\new#1{{#1}}

\def\qrnd{q_{\textup{rnd}}}
\def\qnear{q_{\textup{near}}}
\def\qnew{q_{\textup{new}}}


%% file: tex/intro.tex
\section{Introduction}
\begin{figure*} \centering \subfloat [\sf Tight] {
    \includegraphics[width=0.23\textwidth]{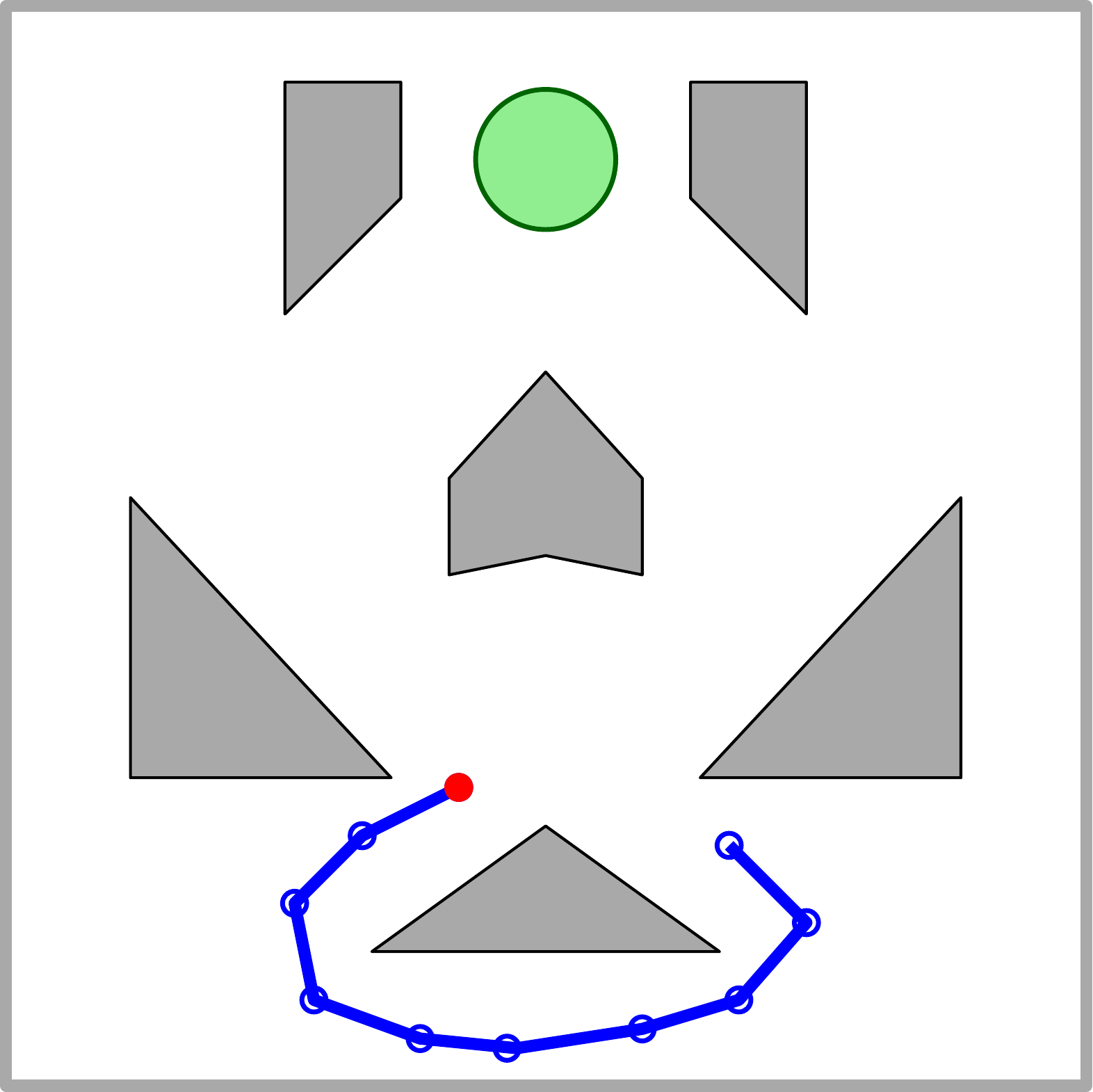}
	\label{fig:tight} } \subfloat [\sf Coiled] {
    \includegraphics[width=0.23\textwidth]{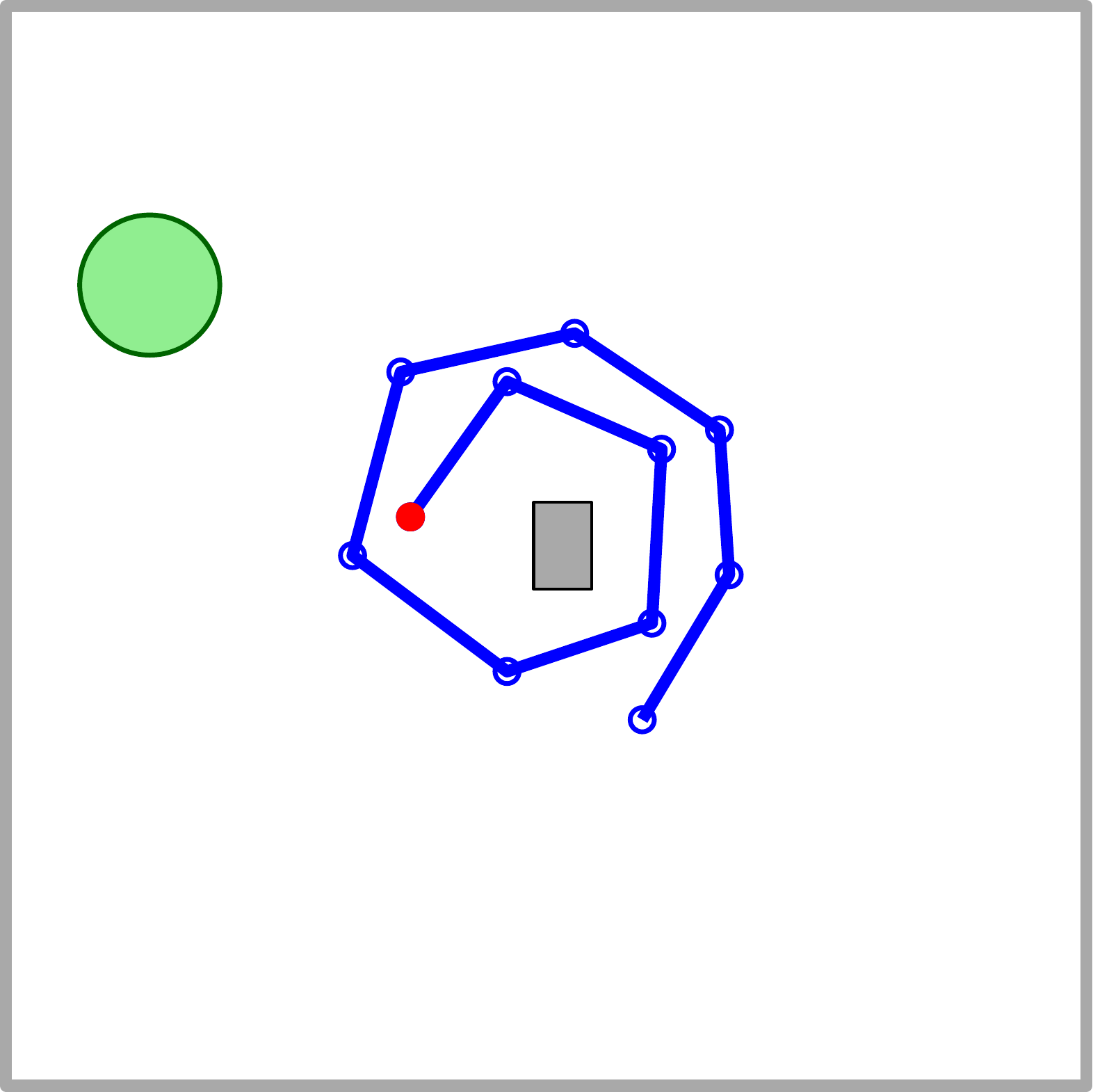}
	\label{fig:coiled} } \subfloat [\sf Bricks (CC)] {
    \includegraphics[width=0.23\textwidth]{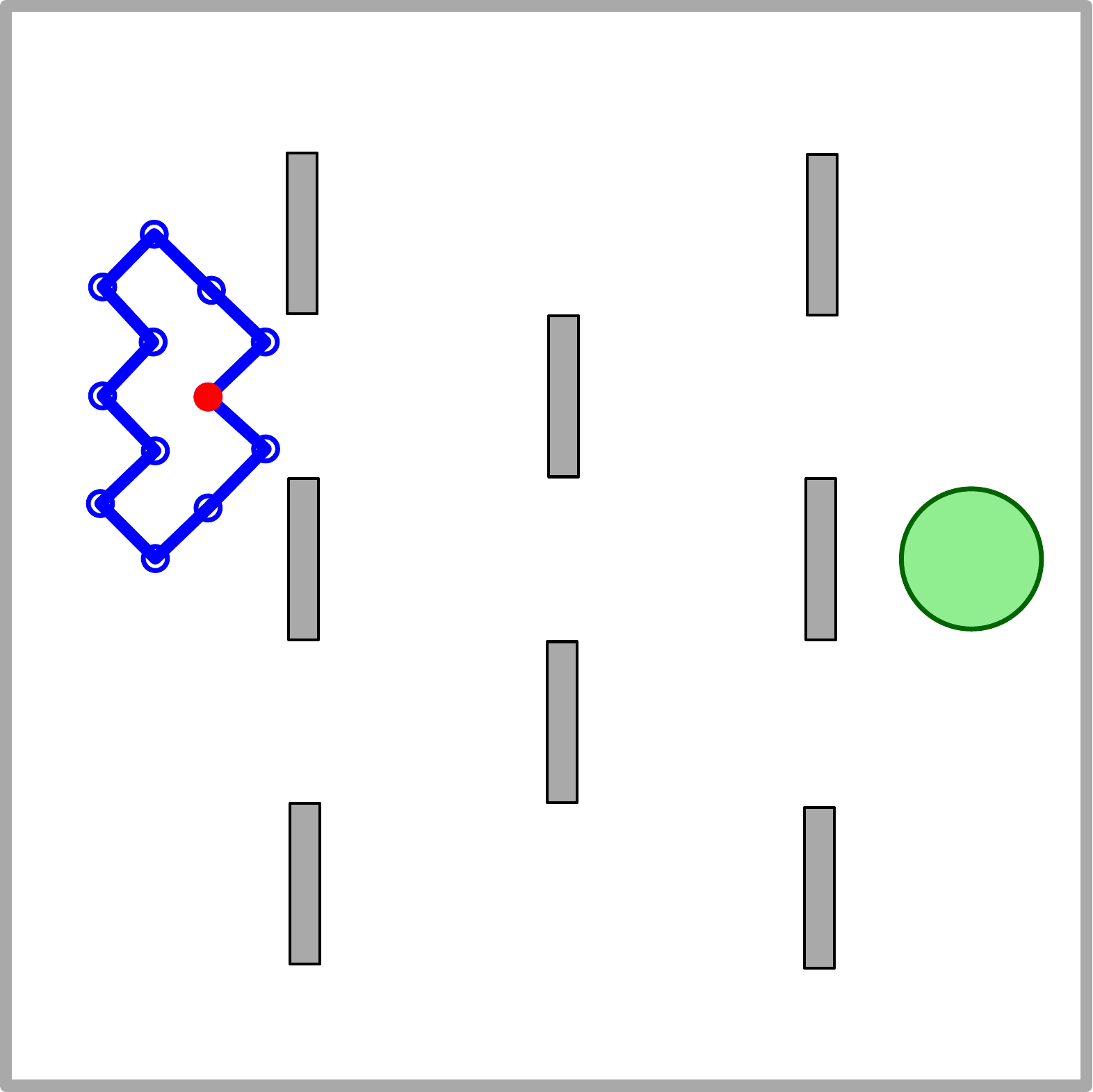}
	\label{fig:bricks} } \subfloat [\sf Gripper] {
    \includegraphics[width=0.23\textwidth]{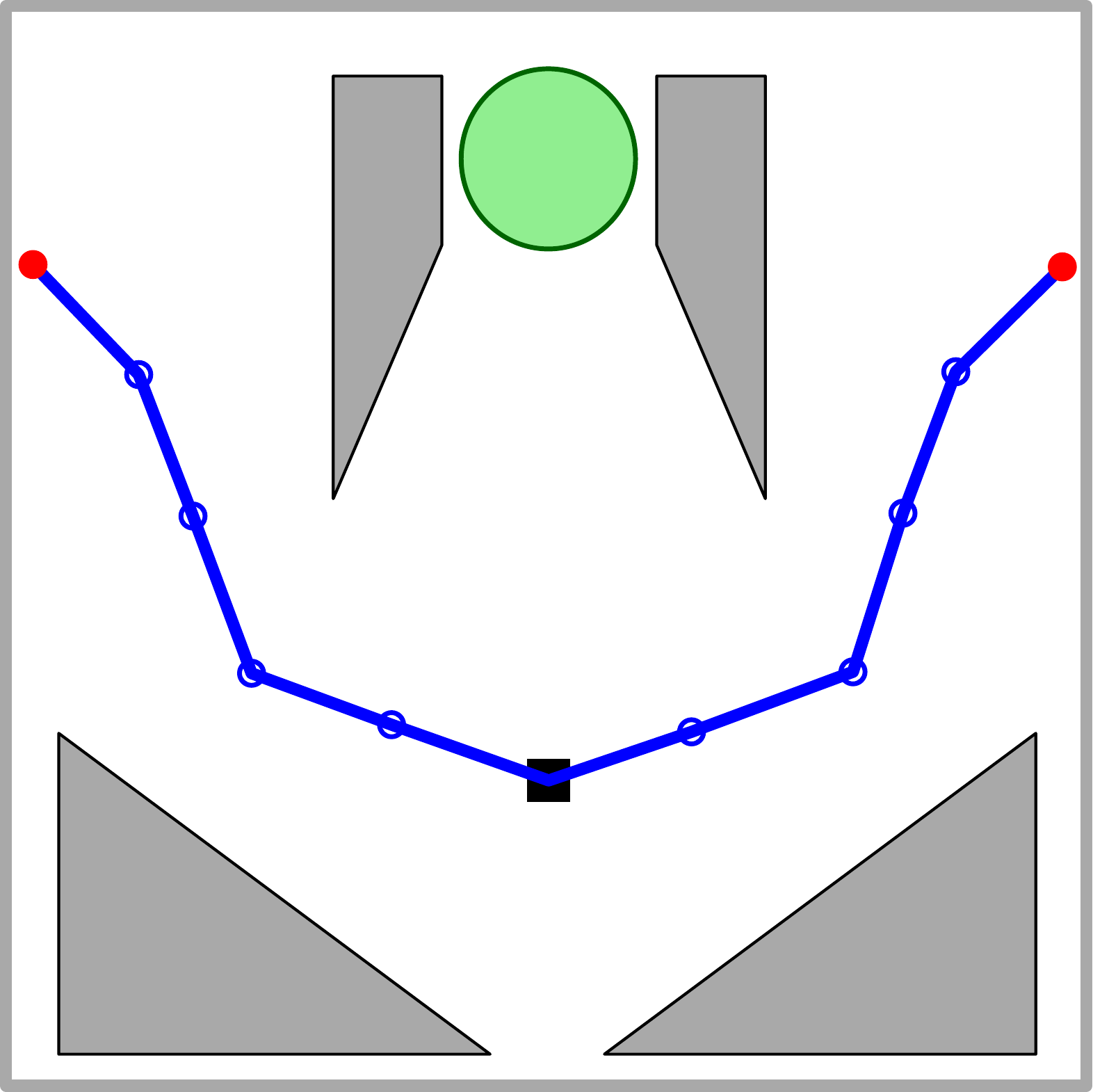}
	\label{fig:gripper} }
  \caption{ Test scenarios.  Robot links and anchor points are depicted in
    solid blue lines and blue circles respectively.  The head of the
    robot (red) needs to move to the target region (green circle)
    while avoiding the obstacles (gray polygons) and self
    intersection.  In (a) and (b) the robot consists of an open link
    chain, whereas in (c) the robot is a closed loop. 
    In (d) the robot's middle joint
    (black square) is permanently anchored to a specific point and
    both endpoints need to reach the target region
  }\label{fig:scenarios}
\end{figure*}
 Motion planning is a fundamental problem in robotics. In its most
 basic form, it is concerned with moving a robot from start to target
 while avoiding collisions with obstacles.  Initial efforts in motion
 planning have focused on designing complete analytical algorithms
 (see, e.g.,~\cite{SS83}), which aim to construct an explicit
 representation of the free space---the set of collision-free
 configurations. However, with the realization that such approaches
 are computationally intractable~\cite{R79}, even for
 relatively-simple settings, the interest of the Robotics community
 has gradually shifted to \emph{sampling-based} techniques for motion
 planning~\cite{CBHKKLT05, L06}.  Such techniques attempt to capture
 the connectivity of the free space by random sampling, and are
 conceptually simple, easy to implement, and remarkably efficient in
 practical settings. As such, they are widely used in
 practice. Another key advantage of these techniques is that
 they are typically described in general terms and can often be
 applied to a wide range of robots and scenarios. However, this also
 has its downsides. Due to the limited reliance of sampling-based
 algorithms on the specific structure of the problem at hand, they tend
 to overlook unique aspects of the problem, which might be exploited
 to increase the efficiency of such methods. For instance, a more
 careful analysis of the specific problem may result in a reduced
 reliance on collision detection, which is often considered to be the
 bottleneck component in sampling-based algorithms.

 In this paper we study the problem of planning the motion of a
 \emph{multi-link} robot, which consists of multiple rigid links
 connected by a set of joints (see Fig.~\ref{fig:scenarios}). We
 assume that the robot is \emph{free flying}, i.e., none of its joints
 are anchored to a specific point in the workspace. We describe a
 novel algorithm which exploits the unique structure of the
 problem. Our work is based on the simple observation that the set of
 configurations for which the robot is not in \emph{self collision} is
 independent of the obstacles or on the exact placement of the robot.
 This allows to eliminate costly self-collision checks in the query
 stage, and to carry them out during the preprocessing stage. The
 novelty comes from the fact that preprocessing needs to be carried
 out once for a given type of robot. This is in contrast to prevalent
 state-of-the-art techniques, such as~PRM*~\cite{KF11}, where the
 preprocessed structure can only be applied to a particular scenario
 and robot type.  In some situations, self-collision
 checks can be as costly as obstacle-collision checks---particularly
 in cases where the robot consists of many links or when the links
 form a closed chain. Moreover, for robots of the latter type,
 computing local paths is particularly costly as the set of
 collision-free configurations lie on a low-dimensional manifold in
 the configuration space.

At the heart of our approach is an implicit representation of the
 \emph{tiling roadmap}, which efficiently represents the space of
 configurations that are self-collision free. In particular, it is
 completely independent of the scenarios in which it can be
 employed. Once a query is given in the form of a scenario---a
 description of the workspace obstacles, a start configuration and a
 target region, the tiling roadmap is traversed using the
 recently-introduced dRRT algorithm~\cite{SSH14}. 

 While our current work deals with free-flying multi-link robots, we
 hope that it will pave the way to the development of similar
 techniques for various types of robots. This may have immediate
 practical implications: when developing a robot for mass production,
 a preprocessed structure, similar to the tiling roadmap, would be
 embedded directly to the hardware of the robot. This has the
 potential to reduce costly self-collision checks when dealing with
 complex robots.

 %

 The rest of the paper is organized as follows: We start by reviewing
 related work in Section~\ref{sec:related_work}, and continue in
 Section~\ref{sec:prelim} with an overview of our technique and some
 preliminary definitions.  In Section~\ref{sec:tiling_roadmap} we
 formally describe the tiling roadmap and in
 Section~\ref{sec:implementation} describe how it should be used in
 order to answer motion-planning queries.  We discuss the properties
 of the tiling roadmap in Section~\ref{sec:coverage} and present
 simulations evaluating our algorithm in Section~\ref{sec:evaluation}.
 Finally, Section~\ref{sec:future} discusses the limitations of our
 work and presents possible future work.

\section{Related work}\label{sec:related_work}
A common approach to plan the motion of multi-link robots is by
sampling-based algorithms \cite{CBHKKLT05, L06}.  While sampling-based
planners such as PRM~\cite{KSLO96} or RRT~\cite{LK01} may be used for
some types of multi-link robots, they are not suited for planning when
the robot is constrained~\cite{YLK01}.  Thus, the recent years have
seen many works attempting to sample valid configurations and to
compute local paths for a variety of multi-link robots differing in
the dimension of the workspace, the type of joints and the constraints
on the system~\cite{H04, HA00, M07, TTCA10, YLK01, ZHL13, HRBV06,
  CS04}.  \new{Additionally, there have been application-specific
  collision-detection algorithms for multi-link robots~\cite{SSL05,
    RCS05, LSHL02, RKLM04} as collision-detection is a key ingredient
  in the implementation of sampling-based motion-planning algorithms.
}

For protein chains, which are typically modelled as high-dimensional
tree-shaped multi-link robots, sampling-based approaches have been
used together with an energy function which guides the search in space
(see, e.g.,~\cite{AS02, LK04, RESH09, SA04}).  Additional applications
of motion planning for multi-link robots are reconfigurable
robots~\cite{KRVM00, NGY00} and digital actor~\cite{Latombe99}.  For
an overview of motion planning and additional applications
see~\cite{L06}.

Specifically relevant to our work is the recently-introduced notion of reachable volumes~\cite{MTA14_ICRA, MTA14_IROS, MTA15}.  The
reachable volume of a multi-link system is the set of points that the
end effector of the system can reach.  The authors show how to compute
the reachable volume and present a method for generating
self-collision free configurations using reachable volumes.  This
method is applicable to open and closed-chain robots, tree-like
robots, and robots that include both loops and open chains. Pan et
al.~\cite{PZM10} introduced a motion-planning algorithm for
articulated models such as multi-link robots, which is integrated in
an RRT-like framework.

Our work shares similarities with the work by Han and
Amato~\cite{HA00}, who studied closed chain system and introduced the
kinematics-based probabilistic-roadmap (KBPRM) planner. This planner
constructs a local PRM roadmap that ignores the obstacles and only
considers the robot's kinematics. Then, copies of the roadmap are
placed in the full configuration space and connections are made
between the copies.  \new{Our work builds on several of the ideas of
  KBPRM.  Specifically, we also exploit the fact that self-collision
  free configurations can be generated while avoiding obstacles and
  copies of these configurations may be placed together with
  connections in the configuration space.}  The main difference from
our work is that the resulting roadmap \new{of KBPRM} depends on a
given workspace environment.  Another difference is that they need to
\new{apply rigid-body local planning} when attempting to connect
copies of the preprocessed local roadmap.  \new{In contrast, the
  implicitly-defined tiling roadmap defined in our work already
  encodes these precomputed self-collision free local plans. Our work
  also bears some resemblance to LazyPRM~\cite{BK00}, which constructs
  a PRM roadmap, but entirely delays collision detection to the query
  stage.  }


\new{Sampling-based algorithms are not the only tool used to address
  the problem at hand.  There have been attempts to study the
  structure of the configuration space of multi-link robots (see,
  e.g.,~\cite{O07, SSB06}) or to explicitly construct it (see,
  e.g.,~\cite{LT05, TM02, SSLT07}).  Space-decomposition techniques
  were used to approximate the structure of the configuration
  space~\cite{PCRT07, PRT09, LuoETAL14} and efficient graph-search
  algorithms were used to search in a configuration space that was
  discretized using a grid~\cite{LGT03}.}

%

%% file: tex/preliminaries.tex
\begin{figure*}
  \centering \subfloat [\sf Base configurations] {
    \includegraphics[height=3.7
    cm]{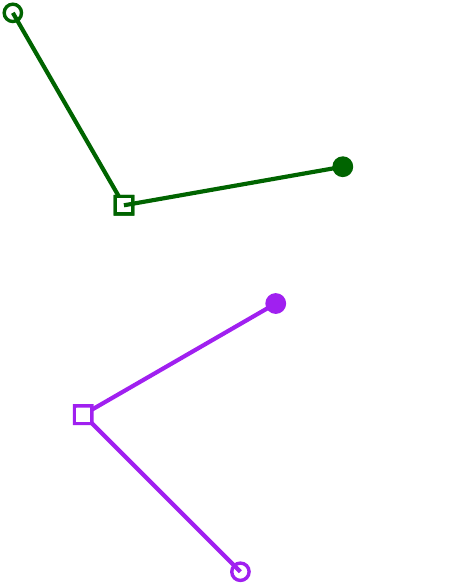}}\quad \subfloat [\sf Translated
  configurations of base roadmaps] {
    \includegraphics[height=4. cm]{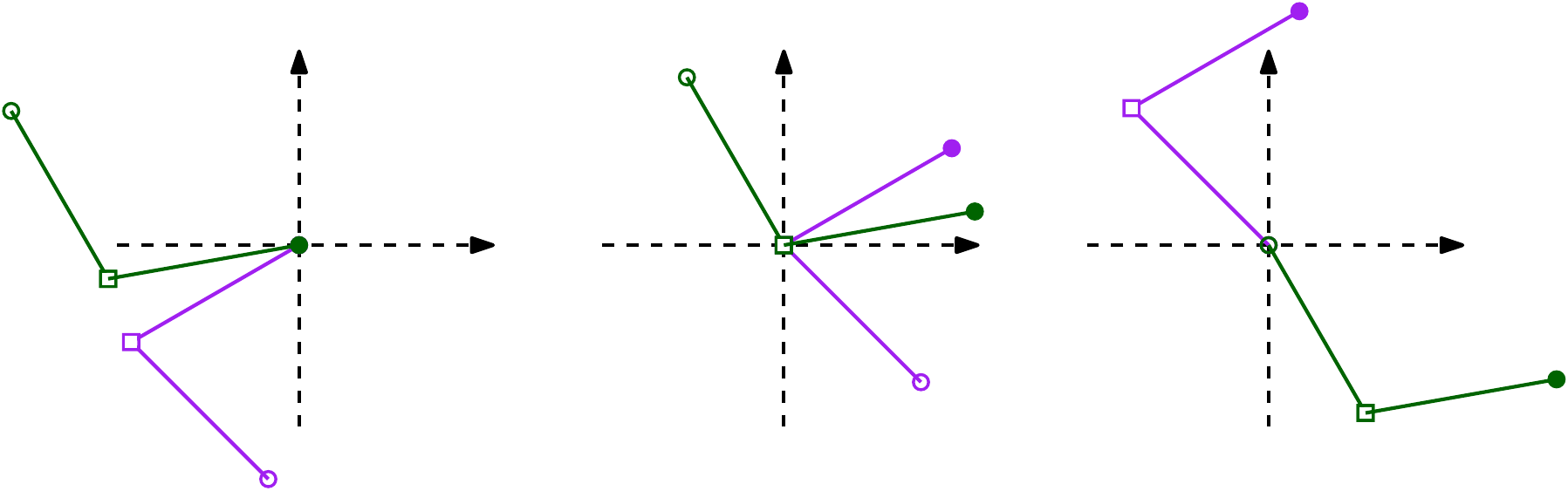}
  }
  \caption{\sf Visualization of base roadmaps for a two-link planar
    robot.  (a)~Two base configurations depicted in green and
    purple. Notice that each anchor point is depicted differently: by
    a circle, a square, or a disc.  (b)~The
    three base roadmaps induced by these two base configurations.}
  \label{fig:visualization}
\end{figure*}

\section{Algorithm overview and terminology}
\label{sec:prelim}
The tiling roadmap~$\G$ is an implicitly-represented infinite graph
that efficiently encodes the space of self-collision free
configurations of a given robot. The structure of $\G$ depends only on
the type of the given robot, and is completely independent of the
workspace scenario in which it can be used.  Every edge of~$\G$
represents a motion in which one of the endpoints of the robot's links
remains fixed in space. We refer to all such endpoints as \emph{anchor
  points}\footnote{The anchor points are not to be confused with the
  robot's joints. A snake-like robot with $m-1$ links has $m-2$ joints
  but $m$ anchor points.}. A motion path induced by $\G$ consists of a
sequence of moves in which the robot alternates between the fixed
anchor points in order to make progress. Given a query, which consists
of a start configuration, a target region, and a workspace
environment, the tiling roadmap~$\G$ is traversed using the
dRRT~\cite{SSH14} pathfinding algorithm (see
Section~\ref{sec:implementation}).  When a configuration or an edge is
considered by the pathfinding algorithm, it is only checked for
collisions with the obstacles.


We now proceed with a set of definitions that will be used in the rest
of the paper. Let $R$ be a multi-link robot moving in some workspace
$\W \subseteq \dR^d$ (where $d \in \{2,3\}$) cluttered with obstacles.
The robot $R$ consists of rigid links and joints connecting them.  
For simplicity of the exposition of the method, we focus here on robots that are ``snake-like'',
i.e., each anchor point connects at most two rotating rigid links and there
are no loops. Thus, we assume that our robot consists of $m-1$ rigid
links and $m$ anchor points. We stress that the technique remains
correct for any other type of a free-flying multi-link robot (see
experiments in Section~\ref{sec:evaluation}).  While a configuration
of such a robot is usually represented by the position $p \in \W$ of
its reference point and the angles of the joints, it will be simpler
to describe our technique while representing a configuration by a
collection of $m$ points in $\dR^d$, which describe the coordinates of
the anchor points.  Thus, we define the configuration space of the
robot to be $\C:=\{(p_1,\ldots,p_m) \ | \ p_i\in \dR^d\}$, such that
the lengths of the links are fixed.  Given an index
$1 \leq j \leq m$ and a point $q \in \dR^d$ we denote by
$\C_{j}(q):=\{(p_1,\ldots,p_m) \ | \ p_j=q, \forall i \neq j \ p_i\in \dR^d \}$, the set
of configurations in which the $j$'th anchor point is $q$.

We denote the \emph{obstacle-collision free space} by
$\F^O\subset\C$. This is the set of configurations in which the robot
does not collide with any obstacle.  In addition, we denote the
\emph{self-collision free space} by $\F^S\subset\C$, which is the set
of configurations in which no two links of the robot
intersect\footnote{%
We consider a configuration to be in self collision if 
(i)~a pair of links that do not share a joint intersect, or 
(ii)~two consecutive links overlap.}.  Finally,
we set $\F:=\F^O\cap \F^S$ and refer to this set as the \emph{free
  space}.  In a similar fashion we define these sets for the case
where the $j$'th anchor point of the robot is fixed at $q\in \dR^d$,
i.e., $\F^O_j(q),\F^S_j(q),\F_j(q)\subset \C_j(q)$.

Given a configuration $C=(c_1,\ldots,c_m)$ and a point $p\in \dR^d$,
let $C+p$ denote the configuration $(c_1+p,\ldots,c_m+p)$.  Namely,
$C+p$ is the configuration obtained by computing a vector sum of each
anchor point with the vector $p$. We say that $C+p$ is the
configuration $C$ \emph{translated} by $p$.  Additionally, for a
configuration $C$, as defined above, and an index $1\leq j\leq m$, let
$j(C):=c_j$.  Namely, $j(C)$ denotes the location of the $j$'th anchor
point of configuration $C$.  Finally, let $\Origin$ denote the origin
of $\dR^d$.


%% file: tex/tiling_roadmap.tex
\begin{figure*}
  \centering \subfloat [\sf ] {
    \includegraphics[width = 0.23 \linewidth]{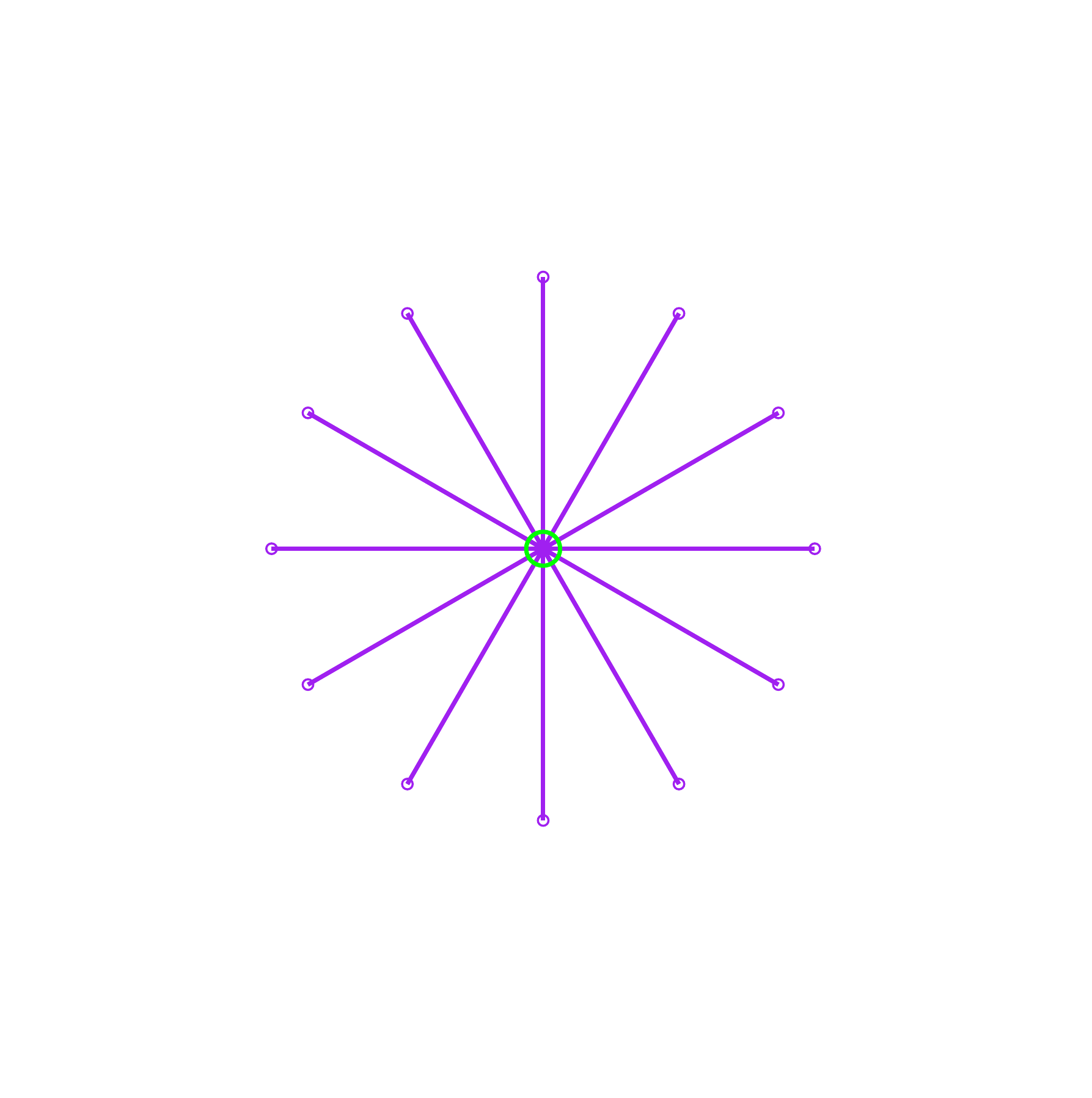}
    \label{fig:dense_1}
  } \subfloat [\sf ] {
    \includegraphics[width = 0.23 \linewidth]{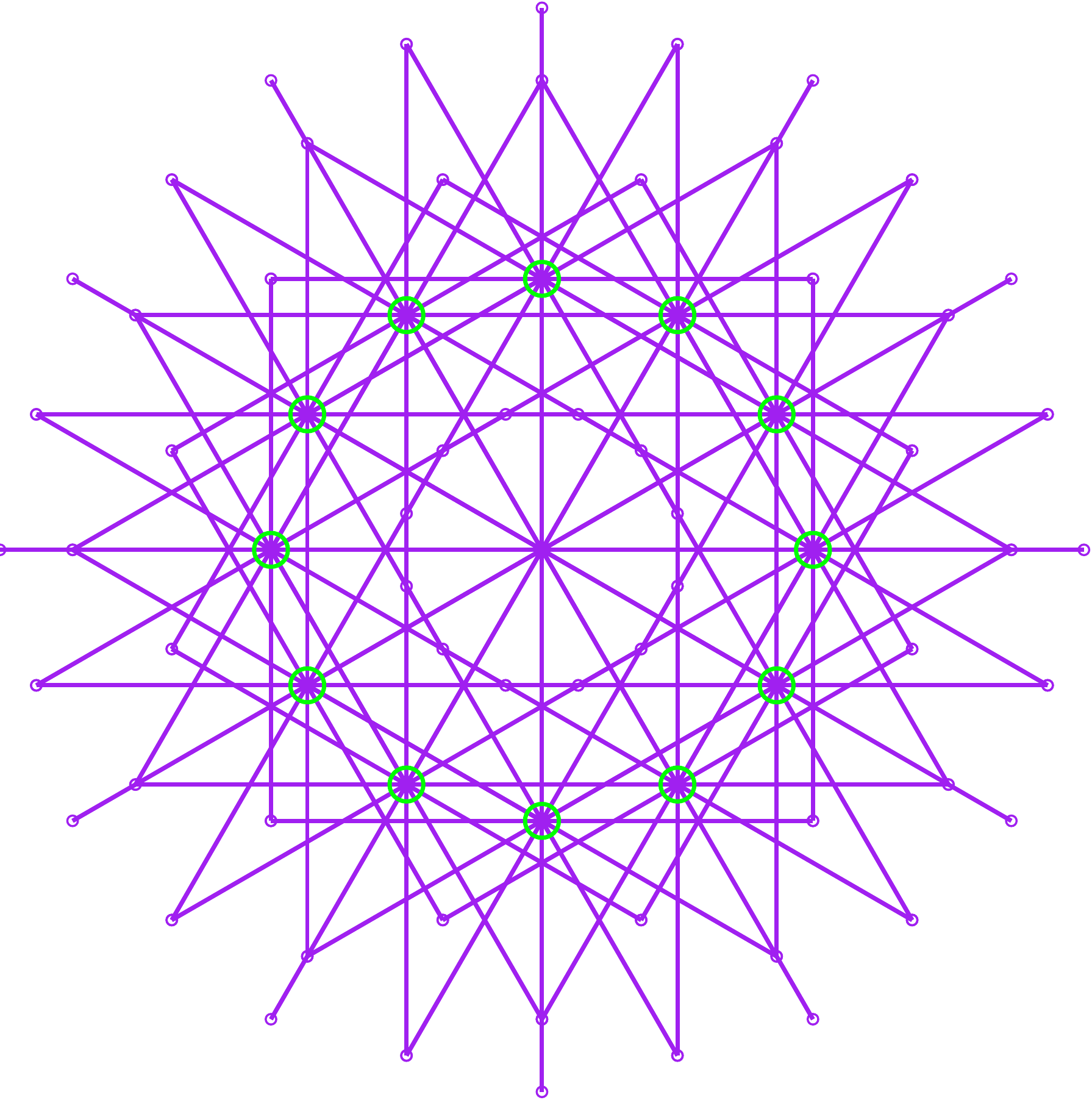}
    \label{fig:dense_2}
  } \subfloat [\sf ] {
    \includegraphics[width = 0.23 \linewidth]{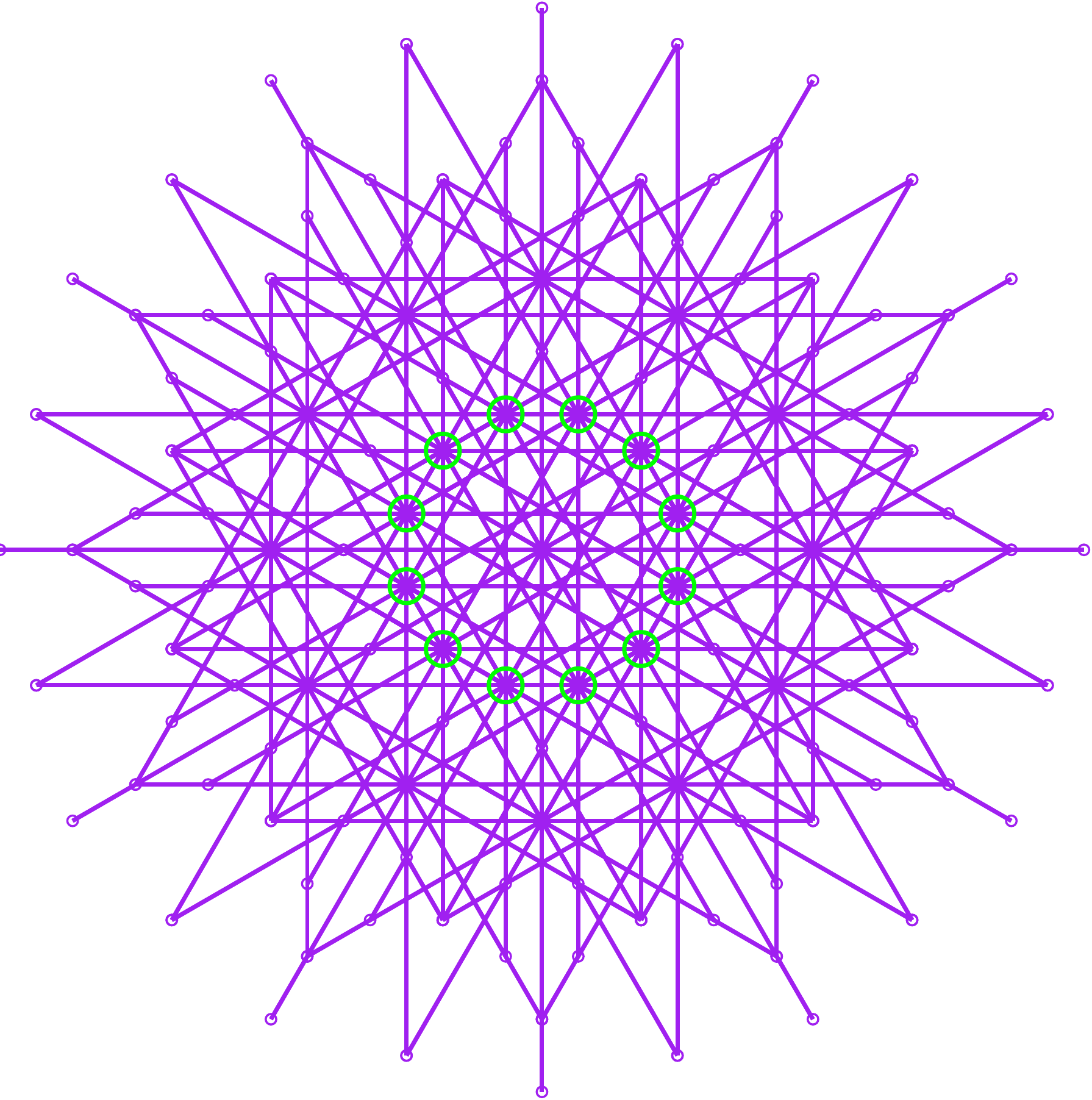}
    \label{fig:dense_3}
  } \subfloat [\sf ] {
    \includegraphics[width = 0.23 \linewidth]{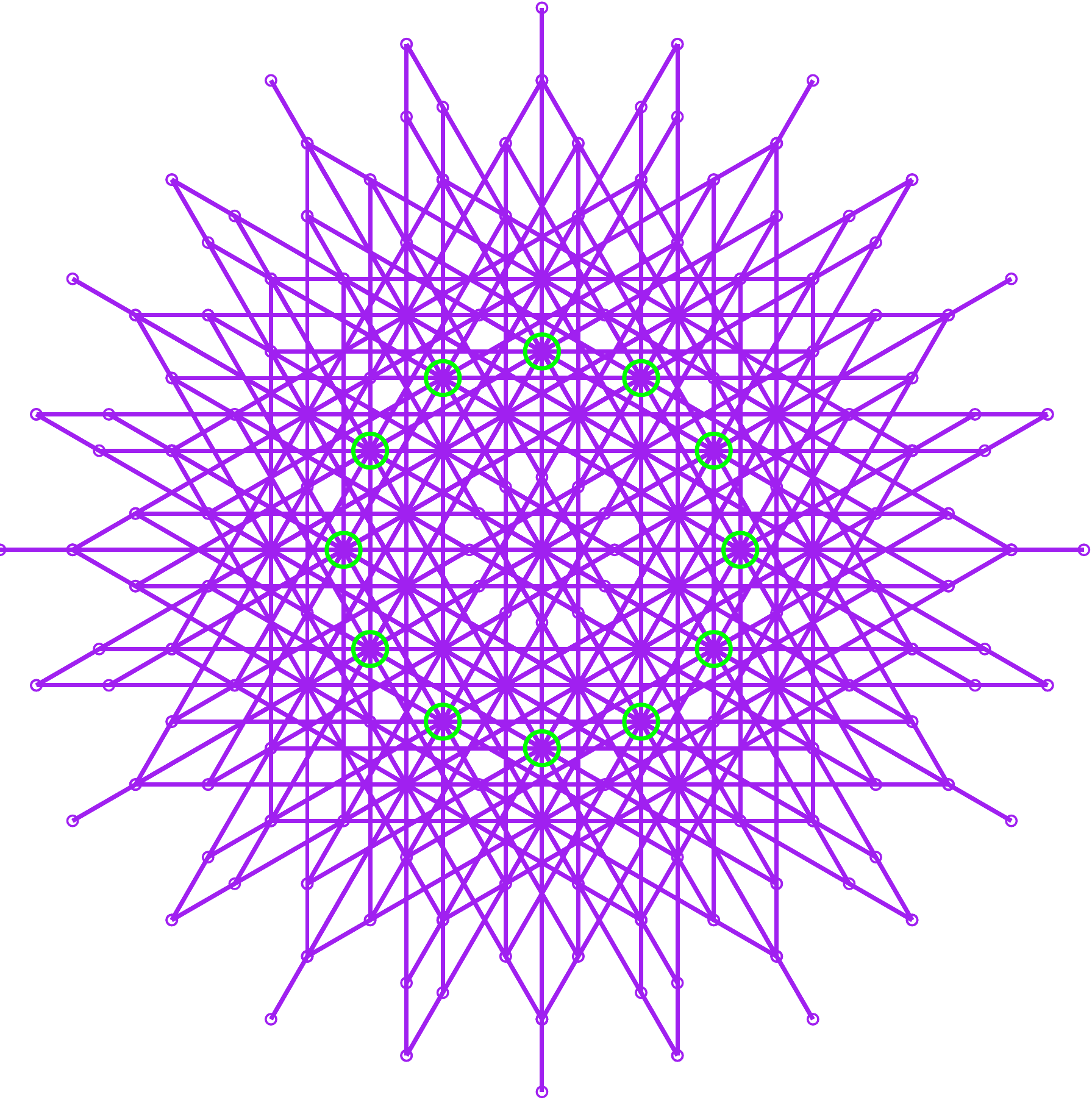}
    \label{fig:dense_4}
  }
  \caption{\sf The tiling roadmap for a toy example which consists of a
    robot with a single link. (a)~Base roadmap for a single-link robot
    with 12 base configurations chosen at fixed intervals of
    $\frac{\pi}{6}$.  (b)-(d)~Iteratively placing the base roadmap on
    endpoints of the link that are closest to the origin (the current
    placements are highlighted by green circles). }
  \label{fig:dense}
\end{figure*}

\section{Tiling roadmaps}
\label{sec:tiling_roadmap}
In this section we formally define the \emph{tiling roadmap}~$\G$,
which approximates the self-collision free space $\F^S$.  We first
describe a basic ingredient of the tiling roadmap called \emph{base
  roadmaps}.  We then explain the role of base roadmaps in the
construction of~$\G$.

\subsection{Base roadmaps for the anchored robot} 
Let $\dC^{\text{base}}:=\{C_1,\ldots,C_n\} \subset \F^S$ be a collection of~$n$
self-collision free configurations called \emph{base
  configurations}\new{, which were uniformly and randomly
  sampled\footnote{In this work we only consider uniform sampling for
    the generation of base configurations. However, we do not exclude
    the possibility that more sophisticated sampling scheme will lead
    to better performance of the framework.}.}  As the name suggests,
the vertices of $\G$ will be based upon the configurations in
$\dC^{\text{base}}$. In particular, every vertex of $\G$ is some translation of a
configuration from $\dC^{\text{base}}$. Conversely, for every $C\in \dC^{\text{base}}$ there exist
infinitely many points $S=\{s_1, s_2,\ldots\}\subset\dR^d$, such that
for every $s\in P$ the configuration $s+C$ is a vertex of~$\G$.

In the next step, we use the configurations in~$\dC^{\text{base}}$ to generate~$m$
roadmaps---one for each anchor point, where the $j$th roadmap
represents a collection of configurations, and paths between them, in
which the robot's $j$th anchor point is fixed at the origin. More
formally, the $j$th roadmap is embedded in $\F_j^S(\Origin)$.  For
each configuration $C_i \in \dC^{\text{base}}$, and every $1\leq j\leq m$, we
consider the configuration $C_{j,i}:=C_i-j(C_i)$ which represents the
configuration $C_i$ translated by $-j(C_i)$.  Clearly, the $j$th
anchor point of $C_{j,i}$ coincides with the origin.  Now, set
$\dC^{\text{base}}_j:=\{C_{j,1},\ldots,C_{j,n}\}$ and note that
$\dC^{\text{base}}_j\subset \F^S_j(\Origin)$. See Fig.~\ref{fig:visualization}.

For every index $j$ we construct the \emph{base roadmap}
$G_j(\Origin)=(V_j(\Origin),E_j(\Origin))$, where
$V_j(\Origin)=\dC^{\text{base}}_j$, which can be viewed as a PRM
roadmap in $\F^S_j(\Origin)$ constructed over the
samples~$V_j(\Origin)$.  We compute for each anchored configuration
$C_{j,i} \in \dC^{\text{base}}_j$ a set of $k$ nearest
neighbors\footnote{To compute a set of nearest neighbors, one needs to
  define a metric over the space. We discuss this issue in
  Section~\ref{sec:evaluation}.} from $\dC^{\text{base}}_j$.  We add
an edge $E_j(\Origin)$ between $C_{j,i}$ and every neighbor if the
respective local path connecting the two configurations is
in~$\F^S_j(\Origin)$.

\subsection{Definition of the tiling roadmap}
So far we have explicitly constructed $m$ base roadmaps
$G_1(\Origin),\ldots,G_m(\Origin)$, where
$G_j(\Origin)\subset \F_j^S(\Origin)$.  We now show that a
configuration that is a vertex in one base roadmap, can also be viewed
as a vertex in the $m-1$ remaining base roadmaps, after they undergo a
certain translation (a different translation for each base
roadmaps). This yields the \emph{tiling roadmap} $\G$, in which
various translations of the base roadmaps are stitched together to
form a covering of~$\F^S$.

Given a point $p \in \dR^d$ we use the notation
$G_j(p) = (V_j(p), E_j(p))$ to represent the roadmap $G_j(\Origin)$
translated by~$p$. We have the following observation, which follows
from the construction of the base roadmaps.
\begin{observation}\label{obs:copies}
  Let $C$ be a vertex of $G_j(\Origin)$ for some $j$. Then~$C$ is also
  a vertex of~$G_{j'}(j'(C))$ for any $j'\neq j$.  Similarly, if $C$
  is a vertex of $G_j(p)$ for some point $p\in \dR^d$, then it is also
  a vertex of $G_{j'}(p+j'(C))$.
\end{observation}

This observation implies that a robot placed in a configuration $C$ of
$G_j(\Origin)$ is not restricted to $\F_j(\Origin)$. In particular, by
viewing $C$ as a vertex of $G_{j'}(p+j'(C))$, it can perform moves in
$\F_{j'}(p+j'(C))$. This argument can be applied recursively, and
implicitly defines the tilling roadmap~\mbox{$\G=(\V,\E)$}. 


We can now proceed to describe the structure of $\G$ in a recursive manner. Initially,
$\G$ contains the vertices of the base roadmaps
$G_1(\Origin),\ldots,G_m(\Origin)$, and the corresponding edges. For
every vertex $C$ of $\G$, and every index $j$, the neighbors of $C$ in
$G_j(j(C))$ are added to $\G$, as well as the respective edges. To
visualize the recursive definition of the tiling roadmap we examine
the simple (self-collision free) case of a robot with a single link
that was preprocessed with $n = 12$ base configurations.  Assume that
for all base configurations, the link's endpoint is fixed at the
origin and the angle of the link with the $x$-axis is chosen at fixed
intervals of $\frac{\pi}{6}$ (see Fig.~\ref{fig:dense_1}).  To avoid
cluttering the figure, we only visualize part of the recursive
construction: We place a copy of this base roadmap on each of the
endpoints of the link (Fig.~\ref{fig:dense_2}) and iteratively repeat
this process for all endpoints around the origin
(Fig.~\ref{fig:dense_3}-\ref{fig:dense_4}). Even for this simple
example with only 12 base configurations we obtain a highly dense
tiling~of~$\F^S$.


%% file: tex/implementation_details.tex

\section{Path planning using tiling
  roadmaps} \label{sec:implementation} Recall that the tilling roadmap
$\G$ represents the self-collision free space $\F^S$ of a given robot.
We describe how~$\G$ is used to find a solution, i.e., a path for the
robot in the fully-free space $\F$, given a query that consists of a
start configuration $S\in \F$, a target region $T\subseteq \F$, and a
workspace environment $\W$.  The solution is found by (i)~adding the
start configuration~$S$ to each base roadmap (together with local
paths in this roadmap) and (ii)~attempting to find a collision-free
path from $S$ to $T$ using~$\G$.  To connect $S$ to each base roadmap
we do the following. Let $S_j:=S-j(S)$ for $1 \leq j \leq m$.  For
every $j$ we connect $S_j$ to $G_j(\Origin)$ by selecting a collection
of nearest neighbors in $G_j(\Origin)$ and applying a \emph{local
  planner} which produces paths in $\F^S_j(\Origin)$. By definition,
$S$ is a vertex of $\G$. It remains to find a path in $\G$ from $S$ to
some other vertex $C\in \V$ such that $C\in T$ using a graph-search
algorithm.  Note that during the search each encountered vertex or
edge of $\G$ should be tested for collision with the obstacles
described by $\W$. Also note that the vertices and edges are
self-collision free by the definition of $\G$.

To search for a path using the tiling roadmap, one may consider
employing standard pathfinding algorithms on graphs such as
A*~\cite{p-hiss}.  However, when the graph is dense, and the problem
lacks a good heuristic function to guide the search, A* (and its many
variants) resort to a BFS-like search which is prohibitively costly
in terms of running time and memory consumption.  This is backed-up by
our experimental work in which A* was unable to make sufficient
progress on~$\G$.

Instead, our motion-planning algorithm integrates the
implicitly-represented tiling roadmap $\G$ with a highly-efficient
pathfinding technique called discrete-RRT~\cite{SSH14} (dRRT).  We
will refer to our framework as \trd, where ``TR'' stands for ``tiling
roadmap''.  dRRT (Algorithm~\ref{alg:drrt}) is an adaptation of the
RRT algorithm for the purpose of exploring discrete,
geometrically-embedded graphs.

\begin{figure} []
  \centering
  \includegraphics[width=0.3\textwidth]{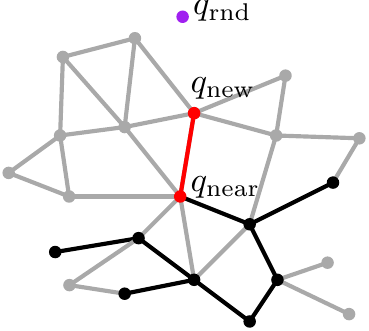}
  \caption{ dRRT algorithm.  The tiling roadmap $\G$ (gray vertices
    and edges) is traversed via subtree $\T$ (explored vertices and
    edges depicted in black).  Extension is performed by sampling a
    random configuration $q_{\textup{rnd}}$ (purple) and locating its
    nearest explored neighbor $q_{\textup{near}}$ in $\T$.  The
    direction oracle returns a neighbor $q_{\textup{new}}$ of
    $q_{\textup{near}}$, which is in the direction of
    $q_{\textup{rnd}}$ (red).  If the edge connecting the two
    configurations is obstacle collision free, $q_{\textup{new}}$ is
    added to the explored tree.  Figure adapted from~\cite{SSH14}.  }
	\label{fig:dRRT}
 \end{figure}

\new{Similarly to its continuous counterpart, dRRT samples a random
  configuration~$\qrnd$ (Alg.~\ref{alg:drrt}, line~2), which is not
  necessarily a vertex of~$\G$. Then, it finds the nearest
  neighbor~$\qnear$ of~$\qrnd$ in the explored portion of~$\G$
  (line~3), which is the tree~$\T$. The difference lies in
  line~4. Whereas RRT usually expands the tree from $\qnear$ towards
  $\qrnd$ by generating a path that linearly interpolates between the
  two configurations of~$\G$, dRRT uses a ``direction oracle'' which
  returns a vertex $\qnew$ such that there is an edge from $\qnear$ to
  $\qnew$ in~$\G$. Moreover, it is guaranteed that the direction
  $\overrightarrow{\qnear \qnew}$ is closest to the direction
  $\overrightarrow{\qnear \qrnd}$ among all the edges of $\G$ incident
  to $\qnear$.
  In the next
  step this edge is checked for obstacle collision (line~5).
  If it is collision free, it is added to $\T$. See a visualization in Fig.~\ref{fig:dRRT}.}
  
  \new{Specifically in \trd,
  the direction oracle is
  implemented in a brute-force manner by going over all the neighbors of
  $\qnear$ in $\G$ and comparing their directions.
  This set of neighbors is extracted as follows:  
  Recall that for every base roadmap $G_j$, there exists a translation $p_j$ such that $\qnear$ is a vertex of $G_j(p_j)$. We take all the neighbors $\qnear$ in $G_j(p_j)$, for every $j$.}

\begin{algorithm}[t]
  \caption{dRRT}
  \label{alg:drrt}
  \begin{algorithmic}[1]
    \LOOP \STATE $\qrnd\leftarrow$
    RANDOM\_SAMPLE( ) \STATE $\qnear\leftarrow$
    NEAREST\_NEIGHBOR($\T,\qrnd$) \STATE
    $\qnew\leftarrow$
    DIRECTION\_ORACLE($\qnear,\qrnd)$ \IF
    {$\qnew \not\in \T$ and VALID($\qnear,\qnew$)} \STATE
    $\T$.add\_vertex($\qnew$) \STATE
    $\T$.add\_edge($\qnear, \qnew$)
    \IF {$\qnew\in T$}
    \RETURN RETRIEVE\_PATH($\T,s,T$)
    \ENDIF
    \ENDIF
    \ENDLOOP
  \end{algorithmic}
\end{algorithm}

 A desirable feature of a sampling-based algorithm is that it
 maintains \emph{probabilistic completeness}.  Namely, as the running
 time of the algorithm increases, the probability that a solution is
 found (assuming one exists) approaches one.  Indeed, the dRRT
 algorithm is probabilistically complete (see~\cite{SSH14}).  We
 believe that, under mild assumptions, one can show that the \trd
 framework is probabilistically complete as well. We discuss this
 issue in the following section.
%


%% file: tex/coverage.tex
\section{Properties of the tiling roadmap}\label{sec:coverage}
We discuss some theoretical properties of the tiling roadmap.
We begin by stating that the tiling roadmap covers the collision-free space
and providing a proof sketch. 
Specifically, we show that for every self-collision free
configuration $C_S\in \F^S$ there exist another configuration $C\in
\V$ such that $C_S$ and $C$ are arbitrarily close as the number of samples tends to infinity. 
We then provide a discussion on the additional steps required to show that our method is probabilistic complete.

\subsection{Coverage}

\begin{figure*}
  \centering \subfloat [\sf ] {
    \includegraphics[width = 3.5cm]{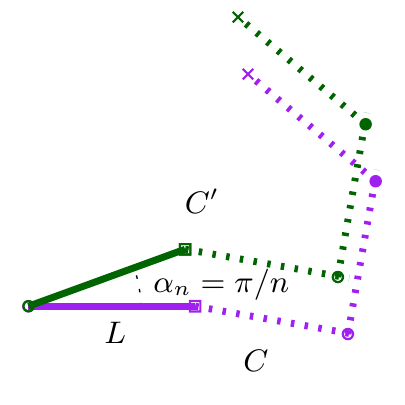}
    \label{fig:coverage1}
  } \subfloat [\sf ] {
    \includegraphics[height = 3.5 cm]{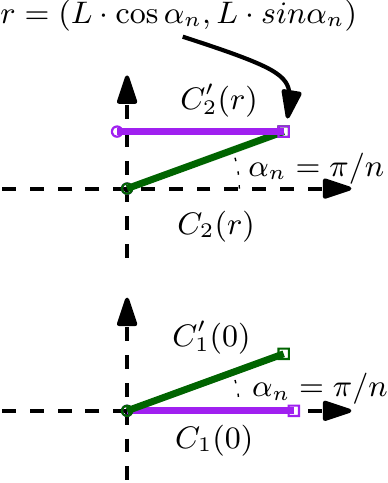}
    \label{fig:coverage2}
  } \subfloat [\sf ] {
    \includegraphics[height = 3.5 cm]{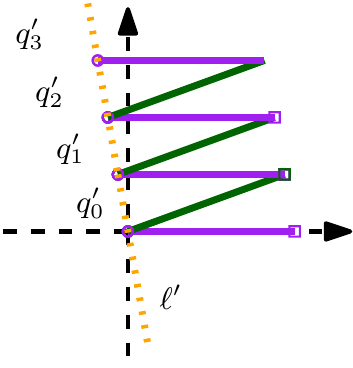}
    \label{fig:coverage3}
  } \subfloat [\sf ] {
    \includegraphics[height = 3.5 cm]{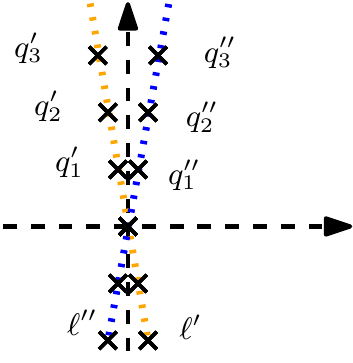}
    \label{fig:coverage4}
  } \subfloat [\sf ] {
    \includegraphics[height = 3.5 cm]{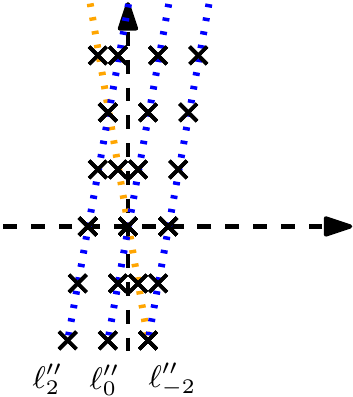}
    \label{fig:coverage4}
  }
  \caption{\sf Construction used head coverage.  
  	(a)~Base configurations $C$ (purple) and $C'$ (green) which are identical
    except for the first angle $\alpha_n = \pi / n$.  The first link 
    (of length $L$)
    of the base configurations is depicted using solid lines and only
    it will be used in~(b) and~(c).  (b)~The two base roadmaps
    $G_1(\Origin), G_2(r)$ for $C,C'$.  (c)~Iteratively constructing the
    points~$\{ q_i' \}$ which densely lie on the line $\ell'$ passing
    through the origin.  (d)~The points~$\{ q_i'' \}$ constructed
    using $C,C''$ ($C''$ is identical to $C$ except for the first
    angle $-\alpha_n$).  (e)~The grid of points (black crosses) where
    the head of the robot may lie (step \textbf{S2}) induced by $\ell'$
    (dashed yellow line) and parallel copies of $\ell''$ (dashed blue line).  
    Only every other copy of $\ell''$ is depicted to avoid cluttering the
    figure.}
  \label{fig:coverage}
\end{figure*}

Intuitively, as the number of base configurations grows, the tiling
roadmap~$\G$ increases its \emph{coverage} of $\F^S$. 
For simplicity, we consider a \emph{planar} snake-like robot 
whose configuration space is represented by
$\dR^2 \times \S^1 \times \ldots \times \S^1$.  We use the term
``head'' to refer to the first anchor point, namely, the first endpoint of
the first link, and denote by $L$ the length of the first link.  We
will use the standard representation of configurations in such a
configuration space.  Namely, we will consider a configuration as a
pair $(p, \Theta)$, where $p = (x,y) \in \dR^2$ is the location of the
head of the robot and $\Theta = (\theta_1 \ldots \theta_{m-1})$ is a
list of angles, where $\theta_i\in \S^1$ is the angle between the
$x$-axis and link $i$. We omit here details regarding the metric in
order to simplify the presentation. Clearly, a rigorous proof will
have to take the specific metric into consideration.

We are now ready to state our claims. First, we show that given a
point $q\in \dR^2$, there is a configuration $C\in \V$, where
$C=(p,\Theta)$, such that $p$ and $q$ are sufficiently close (in the
Euclidean distance). This means that the robot's head can achieve coverage of
$\dR^2$, i.e., \emph{head coverage}. We also require
\emph{angle coverage}: 
given a configuration $C_S=(q,\Theta_S)\in \F^S$,  there exists
$C=(p,\Theta)\in \V$ such that $\Theta$ and $\Theta_S$ are sufficiently
close. Notice that the latter property easily follows from the fact
that base configurations are uniformly sampled from $\F^S$. Combining
the two coverage properties, we have a full coverage of
$\F^S$. 

We show the claim of head coverage using the following two steps. 

\begin{itemize}
\item[\textbf{S1}] Given a start configuration where the head of the
  robot is at the origin $\Origin$, one can construct a straight line
  $\ell'$ in $\dR^2$ such that 
  (i)~$\ell'$ intersects $\Origin$ and 
  (ii)~for every point on $\ell'$, 
  there exists a vertex in the tiling map whose head lies on $\ell'$ 
  and is close to $p$ up to any desired resolution as the number 
  of samples tends to infinity.
\item[\textbf{S2}] The construction in step \textbf{S1} can be used to
  construct two non-parallel lines $\ell', \ell''$ that span $\dR^2$.
  This in turn implies that $\ell', \ell''$ induce a grid 
  such that any point in
  $\dR^2$ is as close as desired to a grid point.
\end{itemize}\vspace{5pt}

\noindent \textbf{Sketch of step \textbf{S1}:}  Set $\alpha_n = \pi / n$ and assume
that the following configurations are in the set of~$n$ base
configurations: $C = ( \Origin, 0, \theta_2 \ldots \theta_{m-1})$ and
$C' = (\Origin, \alpha_n, \theta_2 \ldots \theta_{m-1})$.  Namely, $C$
is the base configuration where the head of the robot is placed at the
origin, the first link lies on the $x$-axis and the remaining angles
$\theta_2 \ldots \theta_{m-1}$ are uniformly chosen.  $C'$ is defined
in a similar manner, except that the first link has a small angle of
$\alpha_n $ with the $x$-axis.  
See Fig.~\ref{fig:coverage1}. 
Moreover, we assume that $C, C'$ are
neighbors in both base roadmaps $G_1(\Origin)$ and $G_2(\Origin)$ 
(namely, the base roadmaps of the first two anchor points).  
Such an assumption is valid since both $C, C'$ are self-collision free and
there exists a self-collision free path between the two.  Thus, we
only assume that they are indeed nearest neighbors, which is true for
sufficiently large value of~$k$, which represents the maximal vertex
degree in every base roadmap. See Fig.~\ref{fig:coverage2}.

We will show that using the assumption that $C, C'$ are neighbors in
both base roadmaps $G_1(\Origin)$ and $G_2(\Origin)$, we can construct a series of
points $\{ q_i' \ | \ i \in \dN \}$ which lie on a line
intersecting the origin. These points represent locations where the head of
the robot may be reached by traversing $\G$ (using exclusively the
configurations $C,C'$ and the base roadmaps $G_1(\Origin),G_2(\Origin)$).  Moreover,
these points get arbitrarily close to one another as the number of
samples grows (which causes $\alpha_n$ to decrease).  The construction
is fairly simple.  Set $q'_0 = \Origin$, $q_1'$ is obtained by moving
from $C_1(\Origin)$ to $C'_1(\Origin)$ in $G_1(\Origin)$ and then
shifting to the roadmap $G_2(r)$ (for the appropriate
point~$r = (L \cos \alpha_n, L \sin \alpha_n)$).  This allows to move from $C'_2(r)$ to
$C_2(r)$ in $G_2(r)$.  Finally, we set $q_1'$ to be the head of the robot
in configuration $C_2(r)$.  This process may be repeated from $q_1'$
to obtain $q_2'$ and henceforth.  Fig.~\ref{fig:coverage3} provides a
visualization of the construction of the points~$\{ q_i' \ | \ i \in \dN \}$.

Recall that $L$ represents the length of the first link.  Using
basic trigonometry we have that
$$ q_i' = ( i \cdot L ( \cos \alpha_n - 1), i \cdot L \sin \alpha_n
).$$
Moreover, all such points $q_i'$ lie on a line $\ell'$ intersecting the
origin and every two consecutive points are of (Euclidean) distance
$$ \Delta(n) = \|q_{i+1}' - q_i' \|_2 = 2 L \sin \frac{\alpha_n }{2} =
2 L \sin \frac{\pi }{2n}.$$
Clearly $\lim_{n \rightarrow \infty} \Delta(n) = 0$.\vspace{5pt}

\noindent \textbf{Sketch of step \textbf{S2}:} The construction of the
line $\ell'$ in step \textbf{S1} was described for a specific pair of
configurations $C, C'$.  Now, if we consider the configuration
$C'' = (\Origin, -\alpha_n, \theta_2 \ldots \theta_{m-1})$, the same
construction holds for the pair of configurations $C, C''$ to obtain
the points $\{ q_i''\ | \ i \in \dN \}$ who all lie on the line $\ell''$ and for which
  $$
  q_i'' = ( i \cdot L ( \cos \alpha_n - 1), - i \cdot L \sin \alpha_n ).
  $$
  Notice that~$\ell'$ and $\ell''$ are not parallel.  
  Moreover, although the  construction of $\ell'$ and $\ell''$
   was shown using the base roadmap
  $G_1(\Origin)$, it holds for any point $p$ where the head of the
  robot may be placed using the tiling roadmap.  Specifically, one can
  construct the set of lines $\{ \ell''_i \ | \ i \in \dN \}$ which are
  parallel to~$\ell''$ such that for every $i \in \dN$, $\ell''_i$ intersects $\ell'$ at the point
  $q'_i$.  This induces a grid on the plane with growing resolution as
  the number of samples grows.  See Fig.~\ref{fig:coverage4}.

\subsection{On the probabilistic completeness of the tiling roadmap}
We hope that the analysis presented for coverage will pave the way to a probabilistic completeness proof of our method.  
Typical proofs of probabilistic
completeness (see, e.g.,~\cite{KKL98, LK02}) rely on two properties of
the examined algorithm. First, they require that the algorithm will
provide \emph{coverage} of the space. Secondly, every two close-by
configurations need to be connected by an edge in the roadmap and the
corresponding local path should remain in the proximity of the
connected configurations. In our case it is unclear whether the tiling
roadmap abides by the second constraint.

%% file: tex/evaluation.tex
\section{Evaluation}
\label{sec:evaluation}
\begin{figure*}[t]
  \centering 
  	\subfloat [\sf Tight] {
    \includegraphics[width=0.48\textwidth]{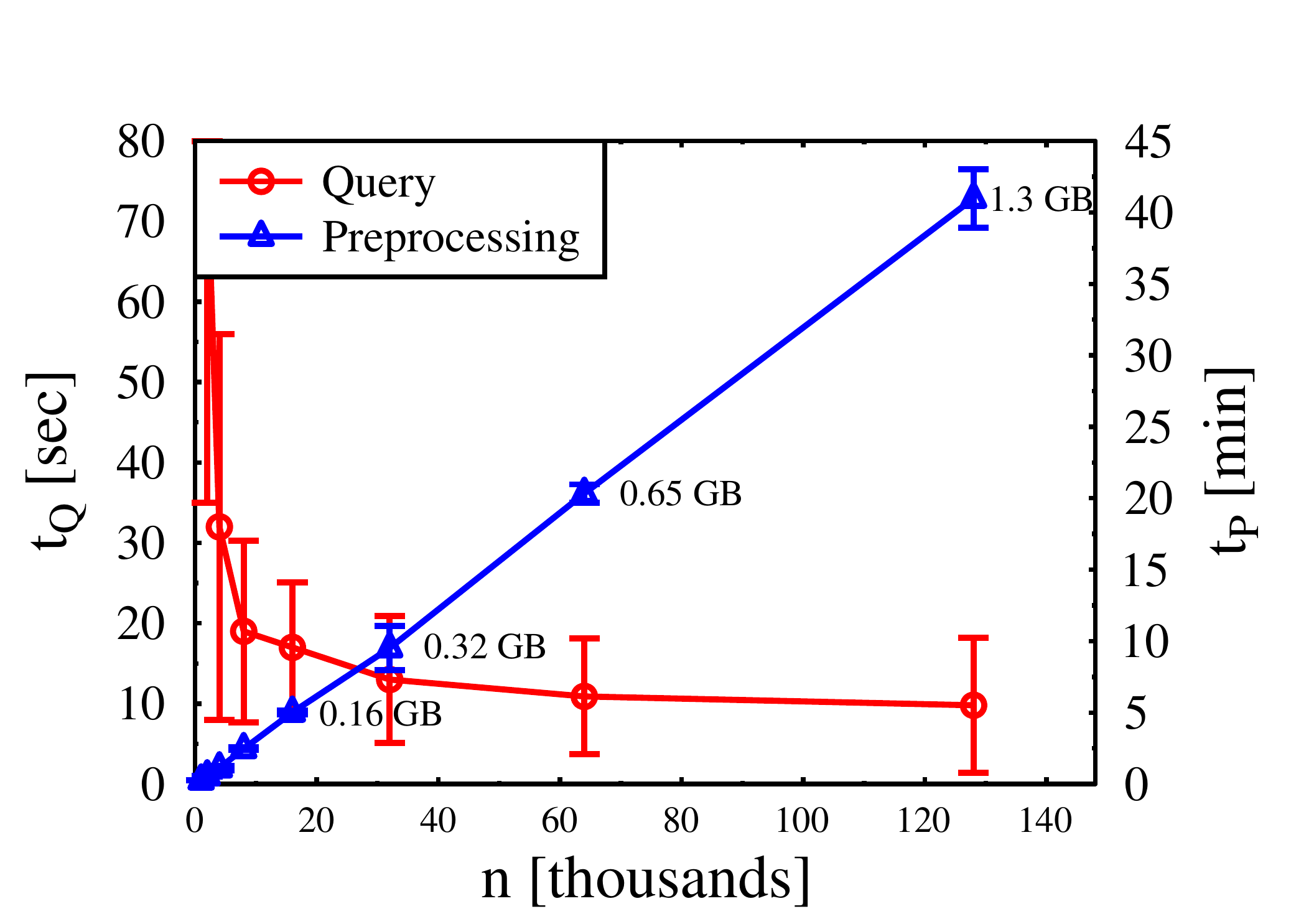}
	\label{fig:times_tight} } 
	\hspace{1mm}
	\subfloat [\sf Bricks CC] {
    \includegraphics[width=0.48\textwidth]{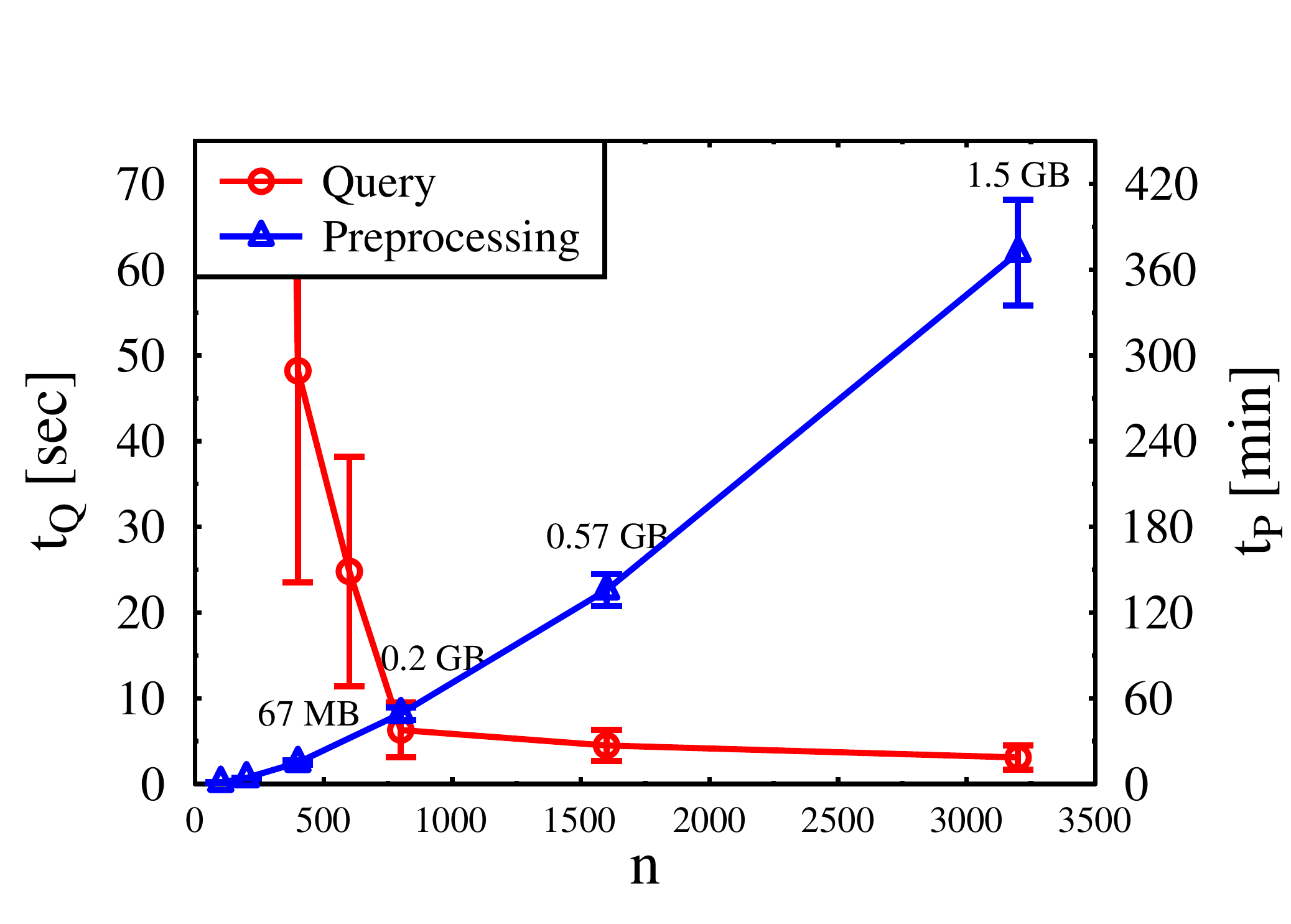}
	\label{fig:times_bricks_cc} }
  \caption{ \new{Quality of results as a function of preprocessing
      times.  Query time $t_Q$ (left y-axis) and preprocessing time
      $t_P$ (right y-axis) as a function of the number $n$ of base
      configurations for the open-chain nine-link robot in the Tight scenario
      (a) and the closed chain robot used in the Bricks scenario (b).
      Error bars denote one standard deviation.  Size of the tiling
      roadmap after preprocessing is added (only partial values are
      given to avoid cluttering the graph).}  }\label{fig:prep}
\end{figure*}
We present simulation results evaluating the performance of our \trd
framework on various scenarios and types of robots in a planar
environment.  We compare its running times with RRT~\cite{kl-rrt} and
RV-RRT~\cite{MTA15}.  Our \Cpp implementation follows the generic
programing paradigm~\cite{A98}, which exploits the similarity between
the behavior of the three algorithms in the query stage, and allows
them to run on a mutual code framework.  The only fundamental
difference lies in the type of the extension method used.  In
particular, whereas RRT and RV-RRT employ a steering function, dRRT
relies on an oracle that can efficiently query for neighbors in the
\emph{precomputed} tiling roadmap.  We use the Euclidean metric for
distance measurement and nearest-neighbor search.  Specifically, every
configuration is treated as a point in $\dR^{2m}$, which represents
the coordinates of the~$m$ anchor points.  Nearest-neighbor search is
performed using FLANN~\cite{muja_flann_2009} and collision detection
is done using a 2D adaptation of PQP~\cite{pqp}.  All experiments were
run on a laptop with 2.8GHz Intel Core i7 processor with 8GB of
memory.

 \subsection{Test scenarios} 
 We experimented with three types of multi-link robots in a planar
 environment  with polygonal obstacles: free-flying open
 chains, free-flying closed chains, and anchored open~chains.

 The first type of robots requires primitive operations (sampling and
 local planning) that are straightforward to implement.  For such
 cases, RRT is the most suitable algorithm to compare with
 \trd. Sample configurations for RRT and \trd were generated in a
 uniform fashion, and local planning was done by selecting one anchor
 point and performing interpolation between the angles of the joints,
 while maintaining the anchor point in a fixed position.  The
 scenarios used to compare the two algorithms for the first type of
 robots are shown in Fig.~\ref{fig:tight}-\ref{fig:bricks}.  In the
 Tight scenario (Fig.~\ref{fig:tight}), a robot with nine links
 navigates in tight quarters among obstacles.  The Coiled scenario
 (Fig.~\ref{fig:coiled}) depicts a robot with ten links which needs to
 uncoil itself. This represents a scenario where the majority of the
 collisions that occur are self collision.  In the Bricks scenario
 (Fig.~\ref{fig:bricks}) a small 13-linked robot needs to move from a
 start configuration with little clearance to the goal.  Note that the
 figure depicts a closed-chain robot as the same scenario is used for
 different robot types. The robot used in the open-chain case is
 identical to the closed chain, except that one of the links is
 removed.

 The second type of robots are free-flying closed chains, which are
 significantly more complex to deal with, as the set of collision-free
 configurations lie on a low-dimensional manifold.
 RV-RRT~\cite{MTA15} is arguably the most suitable algorithm for this
 type of robots. In these settings, \trd uses the primitive operations
 of RV-RRT for sampling and local planning.  We use a robot with 12
 links and evaluate the two algorithms on the Tight
 (Fig~\ref{fig:tight}) and Bricks (Fig~\ref{fig:bricks}) scenarios.
 For \trd the \emph{same} preprocessed roadmap was used to answer the
 queries for the different scenarios, as we use the same robot (see
 Fig.~\ref{fig:scenarios}).

 The third type of robot is an open chain with one of its joints
 permanently anchored to a fixed point in the plane. 
 Here we make a first step toward applying \trd to anchored robot arms.
 In the Gripper scenario (Fig.\ref{fig:gripper}), a 10-link robot is
 anchored at its middle joint and both of the robot's endpoints need
 to reach the goal region.  This simulates two robotic arms that need
 to grasp an object.  We constructed one base roadmap representing
 configurations where the middle joint is anchored.  Currently, it is
 not clear how to extend the tiling-roadmap concept to the case of
 anchored robots (see discussion in Section~\ref{sec:future}).  Thus,
 we resort to dense sampling in the preprocessing stage.  In the query
 stage, this roadmap was traversed using dRRT.  We note that more
 suitable algorithms to solve this problem might exist.  However, the
 simple approach described here, which outperforms RRT, serves as a
 proof of concept for the applicability of our technique to
 anchored-robot settings as well.

\subsection{Experiments}
We first study the affect that the number of sampled base
configurations $n$ has on the query time of \trd. We then proceed to
compare the performance of \trd with RRT and RV-RRT.\vspace{5pt}

\noindent \new{ \textbf{\trd Preprocessing time.} For each robot type,
  we gradually increased~$n$.  For each such value, we constructed the
  tiling roadmap with $k$ nearest neighbors\footnote{We set
    $k= 2e \log n$ in order to reduce the number of parameters
    involved in the experiments. Such a number of neighbors can lead
    to asymptotic optimality of sampling-based motion-planning
    algorithms in certain situations (see,~\cite{KF11}), although we
    do not make such a claim here.}  Finally, each roadmap was used to
  solve the aforementioned scenarios.  Fig.~\ref{fig:prep} reports on
  the results for the open-chain nine-link robot used in the Tight
  scenario and the closed-chain robot used in the Bricks scenario.
  The reported preprocessing times are averaged over five different
  tiling roadmaps and the query times are averaged over 100 different
  queries, for each $n$.  Not surprisingly, as $n$ grows, the
  preprocessing times increase while the query times decrease: As the
  number of base configurations increases, each base roadmap captures
  more accurately the configuration space of the robot anchored to the
  respective joint.  This leads to a better representation of~$\F$
  using $\G$.  Interestingly, for both cases reported, there exists a
  threshold for which the reduction of query times is insignificant
  and comes at the cost of exceedingly-large preprocessing
  times. Similar results were obtained for the other robot types.  }

\noindent \new{ \textbf{Comparison with RRT and RV-RRT.}} In this set
of experiments we compare the query time of \trd with RRT or RV-RRT
for the aforementioned test scenarios. In these experiments \trd
employs the most dense precomputed base roadmaps (see above). We
mention that we are not concerned with the quality of the solution,
and only measure the time required to answer a query successfully.
The preprocessing times for constructing the tiling roadmaps for the
open-chain robots (Fig.~1-3) are fairly low (up to three minutes).
For the Gripper scenario, we had to apply longer preprocessing times
(roughly ten minutes) in order to construct a dense roadmap.  For the
closed-chain scenarios, preprocessing times were in the order of
several hours.  \new{ We note, however, that even fairly moderate
  preprocessing times could have obtained almost the same speedups
  (see Fig.~\ref{fig:prep}).  } For the open-chain robots, both in the
Tight and in the Coiled scenarios, \trd is roughly ten times faster
than the RRT algorithm, while in the Bricks scenario, \trd is roughly
five times faster than the RRT algorithm.  For the Gripper scenario,
\trd is roughly twice as fast as the RRT algorithm.  In more complex
problems with closed-chain robots \trd is roughly more than fifty
times faster than RV-RRT, which is arguably the state-of-the-art for
such problems.

\begin{figure}[t]
  \centering
    \includegraphics[width=0.45\textwidth]{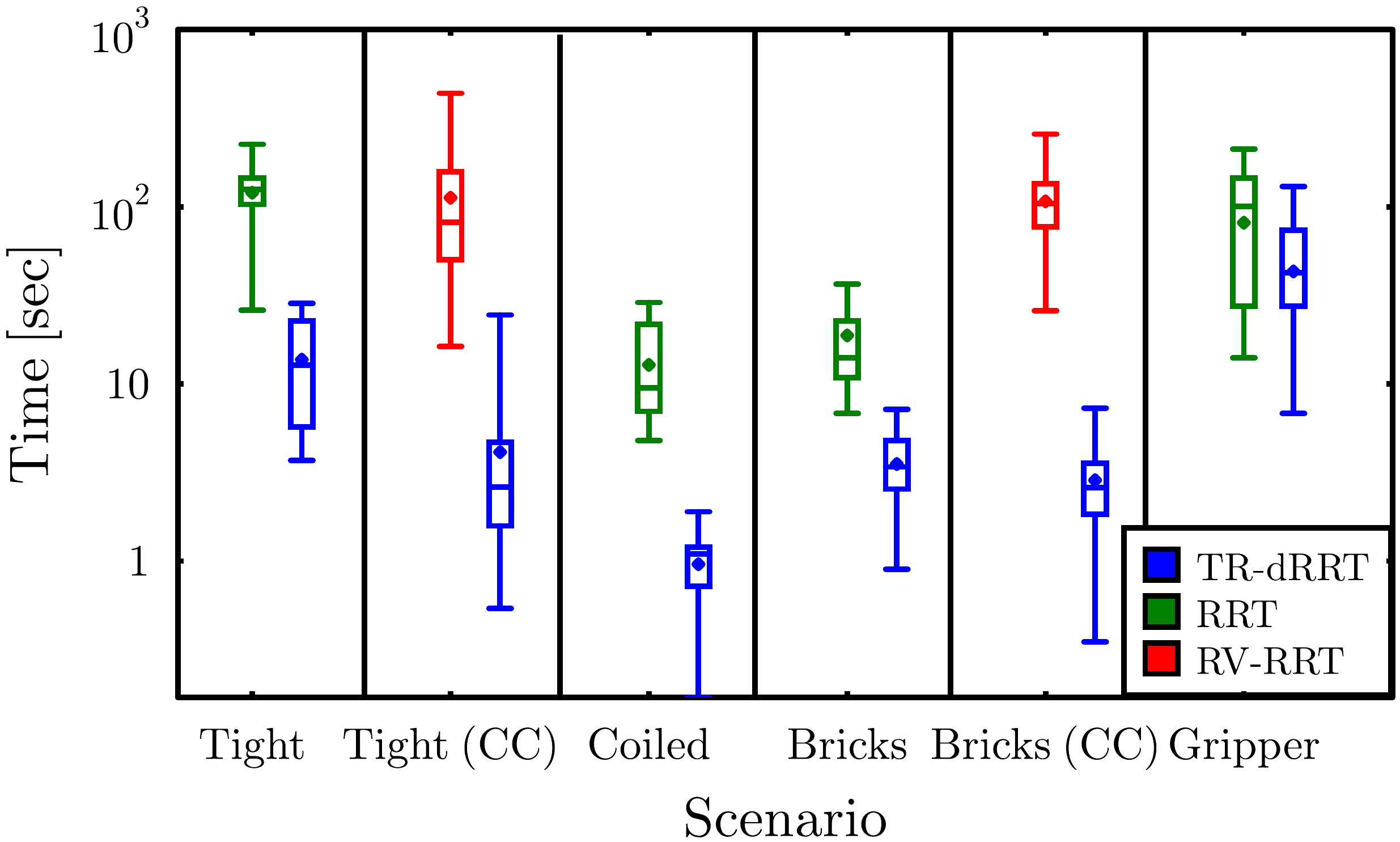}
	\caption{ Running times in seconds for the \trd (blue), RRT
      (green), and RV-RRT (red) algorithms (ten different runs), given
      as box plots: Lower and upper horizontal segments represent
      minimal and maximal running times, respectively; The diamond
      represents the average running time; The line in the middle of
      each box is the median, and the top and bottom of each box
      represent the 75th and 25th percentile, respectively.  CC
      denotes that a closed-chain robot was used in the scenario.
      Notice that the time axis is in log scale.  }
	\label{fig:results}
	\vspace{-5mm}
 \end{figure}


%% file: tex/future_work.tex
\section{Discussion and future work}\label{sec:future}
This paper introduces a new paradigm in sampling-based motion-planning
in which the specific structure of the robot is taken into
consideration to reduce the amount of self-collision checks one has to
perform online. We demonstrate this paradigm using the \trd algorithm
which is designed for free-flying multi-link robots. \trd performs a
preprocessing stage, which results in an implicit tiling roadmap that
represents an infinite set of configurations and transitions which are
entirely self-collision free. Given a query, the search of the
configuration space is restricted to the tiling roadmap. As a result,
no self-collision checks need to be performed, and the query stage is
dedicated exclusively to finding an \emph{obstacle}-collision free
path. 

\new{We note that the preprocessing stage can be performed on stronger
  machinery, e.g., using cloud computing, than the one available to
  the robot in the query stage.  Moreover, the preprocessing stage may
  easily be sped up by computing the individual base roadmaps in
  parallel.}

To conclude, we point out an additional benefit of the TR-dRRT
framework and suggest a direction for future research. \vspace{5pt}

\noindent \textbf{\new{Reducing interpolation costs.}}  \new{Our
  framework can be used not only to eliminate self-collision checks,
  but also to reduce the cost of computing local plans (interpolations
  between two configurations) during the query stage.  The cost of
  computing a local plan becomes significant when the number of
  degrees of freedom of the robot is high or when the robot has closed
  chains.  As the base roadmaps specify which local plans can be used
  in the query stage, such plans can be precomputed and stored for
  every edge of the base roadmaps. Moreover, every local plan can be
  represented in a structure that can reduce the running time of
  obstacle-collision checks. For instance, in our implementation we
  represent every local plan as a polygon that bounds the swept
  area\footnote{The swept area is the overall collection of points
    which the robot covers during a given motion.}  of the respective
  motion. In particular, for every edge of a base roadmap (but not the
  tiling roadmap) we generate the swept area of the respective robot
  motion and generate a polygon which approximates the boundary of the
  swept area. This allows to perform a single obstacle-collision check
  for a given local plan, rather than densely sampling configurations
  along the motion and testing individually each and every
  configuration for collision. Specifically, whenever an edge of the
  tiling roadmap needs to be tested for obstacle collision during the
  query stage, the polygon which corresponds to the base-roadmap edge
  is extracted, translated according to the translation of the tiling
  roadmap edge, and tested for collision with the obstacles.}\vspace{5pt}

\noindent \textbf{Recursive TR-dRRT.}  Although the experimental
results are promising, \trd has its limitations.  The
explicitly-represented base roadmaps should accurately capture the
self-collision free spaces for which one of the anchor points of the
robot is fixed.  For a robot with $D$ degrees of freedom moving in
$\dR^d$, this space is $(D- d)$-dimensional.  Clearly, the favorable
characteristics of our approach diminish as $D$ increases. To overcome
the so-called ``curse of dimensionality'' for this specific type of
robots, we believe that one can apply our technique in a recursive
manner.  For instance, assume that the self-collision free space of a
``snake-like'' robot with $m-1$ links (or $m$ anchor points) can be
captured by a roadmap accurately and efficiently.  Now, given a robot
with $2m-2$ links, it can be decomposed into two parts, consisting of
$m-1$ links each. Then one can generate a tiling roadmap for each of
the two parts and combine the two roadmaps into one, which, in turn,
provides a covering of the entire self-collision free space.  \new{
  This bears resemblance to existing methods which recursively use
  copies of precomputed subspaces in motion-planning
  algorithms~\cite{SHH12} or iteratively solve increasingly difficult
  relaxations of the given motion-planning problem~\cite{BayXieAma05},
  both of which have been shown as effective tools to enhance
  motion-planning algorithms.  }
